\documentclass{article} 
\usepackage{iclr2027_conference,times}

\usepackage{amsmath,amsfonts,bm}

\def\eqref#1{equation~\ref{#1}}

\def\1{\bm{1}}

\DeclareMathAlphabet{\mathsfit}{\encodingdefault}{\sfdefault}{m}{sl}
\SetMathAlphabet{\mathsfit}{bold}{\encodingdefault}{\sfdefault}{bx}{n}

\usepackage{hyperref}
\usepackage{url}
\usepackage{booktabs}
\usepackage{multirow}
\usepackage{float}
\usepackage{caption}
\usepackage{graphicx}
\usepackage{subcaption}
\usepackage{xfrac}
\usepackage{wrapfig}
\usepackage{tabularx}
\usepackage{makecell}
\usepackage{listings}

\title{D-Skill: Distilling Teacher Knowledge through Agent Skills}

\author{%
  Shuo Yuan \quad Dacheng Li \quad Joseph E. Gonzalez \\
  University of California, Berkeley \\
  \texttt{\{shuoyuan, dacheng177, jegonzal\}@berkeley.edu}
}

\iclrfinalcopy 
\begin{document}

\maketitle

\begin{abstract}

Frontier language-model agents excel at complex tasks, but high inference costs make transferring their capabilities to smaller models essential. 
Traditional knowledge distillation achieves this through parameter updates, requiring access to the student’s weights and additional training compute while encoding the transferred knowledge implicitly in the parameters.
We introduce D-Skill, a framework that treats the teacher as a black box and distills its behavioral advantage into an interpretable natural-language skill by analyzing paired rollouts and identifying behavioral divergences and agreements. 
The resulting lessons are consolidated offline, with any compatible skill-update backend, into a skill document that is provided to the student at inference time. D-Skill constructs the skill entirely offline from fixed trajectories, without labels, further environment interaction, or weight updates after rollout collection. Across SpreadsheetBench-Verified, ALFWorld, and chess puzzles spanning multiple teacher–student pairs, D-Skill closes 75\%–95\% of the teacher–student gap without touching model weights. After a single offline pass over the paired traces, D-Skill outperforms one-epoch SFT baselines and remains competitive with multi-epoch SFT. Under the same model family, the resulting student also incurs 5.7×–8.3× lower inference cost per task than the teacher. Our results demonstrate that natural-language skill distillation offers a performant, interpretable, and efficient alternative to weight-based capability transfer.
{\let\thefootnote\relax\footnotetext{Code is available at \url{https://github.com/shuo-yuan/D-Skill}.}}
\end{abstract}

\section{Introduction}
\label{sec:introduction}
\begin{wrapfigure}{r}{0.45\textwidth}
  \vspace{-\intextsep}
  \centering
  \includegraphics[width=0.43\textwidth]{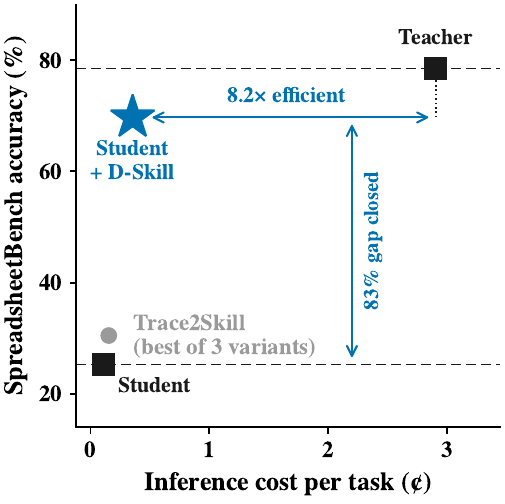}
  \caption{\textbf{D-Skill closes the gap at a fraction of its cost.}
  D-Skill closes 84\% of the teacher-student gap at $8.3\times$ lower inference cost on SpreadsheetBench-Verified. Teacher: GPT-5.6 Terra; student: GPT-5.6 Luna.}
  \label{fig:teaser}
  \vspace{-\intextsep}
\end{wrapfigure}

Frontier agents can reliably solve complex agentic tasks
\citep{openai2026gpt6astra,anthropic2026fable51,jimenez2024swebench,yao2024taubench,xie2024osworld,merrill2026terminalbench}, but their high inference costs \citep{chen2023frugalgpt} make it desirable to transfer their capabilities to cheaper student models \citep{hinton2015distilling,hsieh2023distilling,zeng2023agenttuning,chen2023fireact,chen2024agentflan,kang2025agentdistillation}. Classical knowledge distillation performs this transfer by updating the student's
parameters using supervision such as the teacher's output distribution
\citep{hinton2015distilling,kim2016sequence,gu2024minillm}, complete teacher-generated trajectories \citep{ho2023reasoningteachers,hsieh2023distilling,zeng2023agenttuning,chen2023fireact,chen2024agentflan,kang2025agentdistillation}, or teacher evaluations of student-generated samples
\citep{agarwal2024gkd,bai2022constitutional,lee2024rlaif}. 
Despite differences in supervision, these approaches impose several constraints: weight access is required, excluding many proprietary API-served models; backpropagation adds substantial computational cost; and sufficient training data is needed for the student to generalize. Moreover, the transferred knowledge is encoded in the model's parameters, making it difficult to inspect, revise, or correct after training \citep{meng2022rome,yao2023editing}.

Recent work has shown that \emph{skills}, natural-language artifacts encoding reusable task knowledge, strategies, and procedures, can substantially improve agent performance without modifying model weights \citep{anthropic2025agentskills,ni2026trace2skill,yang2026skillopt,zhang2025ace}.
Because skills remain external to the model, they are portable across deployments and can be inspected or edited after construction \citep{anthropic2025agentskills}. 
Existing skill-generation methods are generally not organized around the teacher--student relationship central to distillation. Teacher-driven methods compile stronger-model trajectories or teacher-generated modules without conditioning the transfer on the target student's behavior \citep{ni2026trace2skill,fu2024autoguide,qiu2025agentdistill}. Student-centered methods instead learn from the target agent's own experience, including contrasts between successful and failed rollouts, without a matched stronger-agent reference \citep{shinn2023reflexion,zhao2024expel,wang2024awm,suzgun2025cheatsheet,zhang2025ace,yang2026skillopt,ma2026skillgen}. D-Skill instead constructs a student-targeted skill by comparing teacher and student trajectories collected on the same tasks. We ask whether their behavioral differences can be distilled, in a single offline pass over fixed traces, into an external textual update.

We propose \textbf{D-Skill}, a method for distilling model capabilities into a skill by comparing paired teacher and student rollouts. For each unlabeled task, we obtain one rollout from each model, either through fresh execution or from previously saved traces. A difference analyst extracts lessons from both divergences and agreements between the two behaviors. Divergences identify states in which the student behaves differently from the teacher and pair the student's behavior with the teacher's alternative, while agreements identify shared strategies that should be preserved. These lessons are consolidated into a single skill, which is loaded into the student's context at inference time. Once the rollouts have been collected, D-Skill requires no answer keys, labels, held-out evaluation, or further environment interaction. The teacher and student contribute only through their fixed traces: neither is queried or fine-tuned during analysis and consolidation, and no model weights are modified.

We evaluate D-Skill on SpreadsheetBench Verified, ALFWorld and chess puzzles using different teacher-student pairs with different model families. As presented in section \ref{main_result}, across all three benchmarks and different teacher-student settings, D-Skill closes \(83.4\%\) to \(89.3\%\) of the teacher--student performance gap on SpreadsheetBench Verified, \(84.6\%\) to \(94.6\%\) on ALFWorld, and \(75.3\%\) to \(87.7\%\) on chess puzzles. This transfer also yields substantial inference-cost savings: in the GPT-5.6 Terra–Luna setting, the D-Skill student costs $8.3\times$ less per task than the teacher on SpreadsheetBench Verified and $5.7\times$ less per episode on ALFWorld (see section \ref{subsec:cost}). We further compare D-Skill with supervised fine-tuning and find that it provides stronger transfer after a single pass over the teacher corpus while attaining performance comparable to multi-epoch fine-tuning (Section~\ref{subsec:sft}). These results demonstrate that textual skill distillation can match weight-based transfer in the evaluated setting while producing an artifact that remains readable, editable, and portable. In summary, this paper makes the following contributions:

\begin{itemize}
    \item We formulate \emph{skill distillation} as an alternative to weight-based knowledge distillation, encoding transferred capabilities in an external, interpretable artifact that requires neither weight access nor gradient-based training.

    \item We introduce \textbf{D-Skill}, a simple yet effective method that distills paired teacher and student rollouts by identifying their behavioral divergences and agreements and consolidating the resulting lessons into a reusable skill. After rollout collection, the method requires no labels, held-out evaluation, or further environment interaction.

     \item We demonstrate across three benchmarks and multiple teacher--student pairs that D-Skill closes most of the performance gap. In addition, it also outperforms the evaluated skill-learning baselines while retaining substantially lower inference cost than the teacher.
\end{itemize}
\section{Related Work}
\label{sec:related_work}

%
\paragraph{Knowledge distillation.}
Knowledge distillation trains a compact student to match a stronger teacher's
predictions \citep{bucilua2006model,hinton2015distilling}, extended to sequence-level
imitation \citep{kim2016sequence}, compressed language models \citep{sanh2019distilbert},
and on-policy variants that supervise the student's own rollouts to close the
train--inference mismatch \citep{gu2024minillm,agarwal2024gkd,xu2024speculative,zhao2026opsd}.
Throughout, two assumptions persist: the transfer \emph{medium} is parameters
(gradient updates, so closed-weight students are out of reach), and the signal
is distributional (logits or token-level divergences, requiring white-box teacher
access and a shared tokenizer). D-Skill relaxes both: knowledge is retained as an
explicit textual skill consumed in context by black-box students, and the signal is the behavioral contrast between teacher and student trajectories on shared tasks.

\paragraph{Skill and memory-augmented agents.}\label{sec:rw-skills}
A complementary line of work equips agents with persistent memories or skills---external artifacts carried across tasks and consulted at inference time. Reflexion stores reflections on past failures \citep{shinn2023reflexion}, Voyager builds libraries of executable skills through exploration \citep{wang2023voyager}, and subsequent work consolidates trajectories into procedural memory \citep{wang2024awm,fang2025memp,li2026rlem}. SkillGen contrastively induces and validates skills from an agent's successful and failed rollouts \citep{ma2026skillgen}. Teacher-driven approaches instead transfer stronger capabilities: Trace2Skill compiles stronger-model trajectories into textual skills \citep{ni2026trace2skill}, while AgentDistill packages teacher-generated procedures as reusable MCP modules \citep{qiu2025agentdistill}. 
D-Skill instead constructs a student-targeted skill by directly comparing teacher and student trajectories. 
D-Skill addresses this gap by contrasting teacher and student trajectories, producing guidance targeted to their behavioral differences.

\paragraph{Automatic textual optimization.}\label{sec:rw-texopt}
A third line improves a text artifact through iterative textual updates: APE and OPRO
search over instructions with LLM proposers \citep{zhou2023ape,yang2024opro},
Promptbreeder evolves prompts self-referentially \citep{fernando2023promptbreeder},
DSPy compiles multi-stage pipelines by tuning their prompts against a validation metric
\citep{khattab2024dspy}, and TextGrad backpropagates natural-language critiques as
``textual gradients'' \citep{yuksekgonul2024textgrad}. The same machinery has recently
been applied beyond instructions: GEPA evolves prompts by reflecting on execution
traces \citep{agrawal2025gepa}, and SkillOpt optimizes a portable skill file through
gated patch-style edits validated on held-out tasks \citep{yang2026skillopt}, one
instance of a growing family of self-evolving skill methods \citep{jiang2026sok}.
These methods treat text as an optimizable artifact, typically using task evaluation
to select or refine updates. D-Skill follows the same broad paradigm but focuses on
teacher--student distillation: it derives textual updates from behavioral contrasts
between a stronger teacher and a target student on matched tasks. D-Skill thus extends
automatic textual optimization to capability transfer while remaining compatible with
different skill-update backends.

\section{Method}
\label{sec:method}
\begin{figure}[t]
    \centering
    \includegraphics[width=\linewidth]{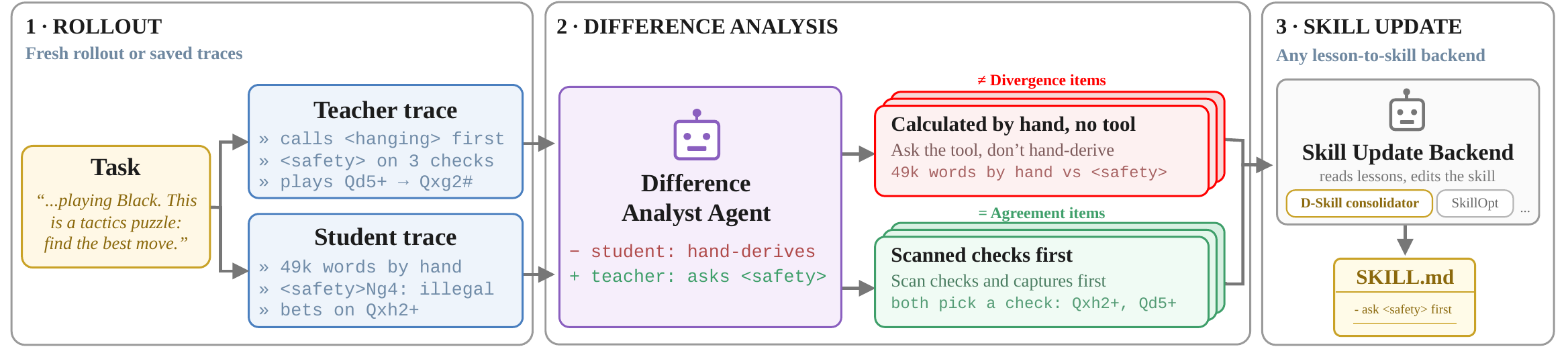}
    \caption{
    \textbf{Ground-truth-free skill distillation from paired rollout traces.}
    (1) \textbf{Rollout:} Paired teacher and student traces are obtained
    from fresh rollouts or saved executions.
    (2) \textbf{Difference analysis:} A Difference Analyst Agent compares
    the fixed traces without answer keys, labels, or further
    environment interaction, extracting divergence and
    agreement items.
    (3) \textbf{Skill update:} A Consolidation Agent converts the lessons
    into patches, merges them, and applies the update to \texttt{SKILL.md}.
    }
    \label{fig:paired-rollout-pipeline}
\end{figure}

Figure~\ref{fig:paired-rollout-pipeline} summarizes D-Skill, which consists
of three stages: rollout collection, difference analysis, and skill update.
No stage computes a gradient or requires access to the parameters of either
model. Collecting fresh rollouts requires the teacher and student to interact
with the task environment; when saved traces are available, this interaction
can instead have occurred during an earlier execution. In either case, all
subsequent analysis and skill construction operate offline on fixed traces.

\subsection{Problem Setup}
Let $\mathcal{T}=\{t_1,\ldots,t_n\}$ be a set of unlabeled tasks sampled from
a task distribution $\mathcal{D}$, and let $\pi_S$ and $\pi_T$ denote a
student and a stronger teacher policy operating in the same task and action
space. Given paired trajectories
$\{(\tau_S(t),\tau_T(t)):t\in\mathcal{T}\}$, our objective is to construct a
natural-language skill $s$ such that the skill-conditioned student $\pi_S^s$
improves over the original student without modifying its parameters. If
$J(\pi)$ denotes expected task performance, we quantify the transfer by the
fraction of the teacher--student gap closed:
\[
    \operatorname{GapClosed}(s)
    =
    \frac{J(\pi_S^s)-J(\pi_S)}
         {J(\pi_T)-J(\pi_S)}.
\]

\subsection{Rollout Collection}

For each task $t\in\mathcal{T}$, we collect one trajectory from each model,
$\tau_S(t)$ and $\tau_T(t)$, recording the actions issued and the observations
returned. The paired trajectories are the unit of supervision; no later stage analyzes either trajectory in isolation. 
The two policies attempt each task independently, with neither conditioned on the other’s behavior. When trajectories are collected specifically for D-Skill, both policies interact with the task environment through ordinary execution; alternatively, trajectories may be retrieved from logs of prior executions, allowing distillation to begin directly from saved traces. The two trajectories need only correspond to the same task and use a shared action interface. Once the trajectories have been collected, D-Skill operates fully offline: neither \(\pi_S\) nor \(\pi_T\) is invoked again, and skill construction requires no further environment access, ground-truth labels, rewards, or evaluation feedback.

\subsection{Difference Analysis}

In the second stage, a difference analyst agent $\mathcal{A}$ converts each
paired student and teacher trace into a set of structured, reusable
lessons. The analyst can inspect the actions and observations recorded in
the traces, but it has no access to answer keys, labels, or additional environment interaction.

Rather than asking the analyst to compare the trajectories freely, we require it to examine each pair along a predefined set of behavioral dimensions. These dimensions cover, for example, what information each agent gathered before acting, how it selected and executed actions, how it responded to errors or unproductive intermediate observations, and how it verified the final result. This shared schema makes lessons comparable across task pairs and discourages the analyst from emphasizing superficial differences in naming, phrasing, or reasoning style. Each lesson must cite concrete evidence from the paired traces and express a task-general principle rather than an instance-specific correction. Details about the prompt for the analyst can be found in appendix \ref{app:analyst-prompts}.

For each trajectory pair, the analyst produces at most three lessons, ranked by their anticipated usefulness on future tasks, and assigns each to one of two classes: a \emph{divergence lesson} identifies a behavioral difference between the two agents, specifying the student’s behavior, the teacher’s alternative, and a reusable instruction intended to address the gap; an \emph{agreement lesson} identifies a behavior shared by both agents that is judged worth preserving in the distilled skill.

\subsection{Skill Update}

To construct the final skill document, we use a modified version of the
Trace2Skill consolidation backend \citep{ni2026trace2skill} as the D-Skill consolidator. We retain its map--reduce--apply procedure but adapt its input to consume the structured
divergence and agreement lessons produced by our difference-analysis
stage. The complete lesson pool is shuffled and partitioned into $k$ chunks, which are
processed sequentially to update the skill document.
For each chunk, a consolidation agent $\mathcal{C}$ performs three phases.
In the map phase, it independently converts each lesson into a
candidate patch to the current skill, such as appending to an existing
section, replacing existing content, or creating a new section. In the
reduce phase, it merges the patches by removing duplicates,
resolving conflicts, and generalizing related rules. Finally, in the
apply phase, the merged patch is deterministically incorporated into
the current skill before the next chunk is processed.

We use this adapted Trace2Skill backend throughout our experiments. D-Skill's
output from the preceding stage is a general pool of structured lessons, so
other compatible skill-updating backends could be substituted without
changing the rollout-collection or difference-analysis stages, more details can be found in section \ref{backends}.
\section{Experiments}
\label{sec:experiments}

\providecommand{\err}[1]{{\scriptsize$\pm$#1}}

\begin{table}[t]
\centering
\caption{Main results. Each block reports the bare student (floor), the student with the distilled D-Skill skill, and the teacher; \%GC is the fraction of the
teacher--student gap closed.}
\label{tab:main}
\footnotesize

\begin{subtable}{\linewidth}
\centering
\subcaption{SpreadsheetBench-Verified (SSB-V) and ALFWorld}
\label{tab:main-a}
\setlength{\tabcolsep}{3.5pt}
\begin{tabular}{llcccc}
\toprule
& & \multicolumn{2}{c}{SSB-V} & \multicolumn{2}{c}{ALFWorld} \\
\cmidrule(lr){3-4}\cmidrule(lr){5-6}
Teacher & Student config & Score & \%GC & Score & \%GC \\
\midrule
\multirow{5}{*}{\shortstack[l]{DeepSeek-\\V4-Pro-0423 \\+ Skill}}
 & Qwen3.6-35B-A3B      & $28.0$\err{1.4} & ---  & $59.0$\err{0.7} & ---  \\
 & \quad + D-Skill          & $\mathbf{71.7}$\err{3.6} & $\mathbf{85.1\%}$ & $\mathbf{84.2}$\err{3.3} & $\mathbf{85.7\%}$ \\
 & GPT-5.6 Luna     & $25.3$\err{1.1} & ---  & $55.2$\err{4.2} & ---  \\
 & \quad + D-Skill          & $\mathbf{73.3}$\err{1.0} & $\mathbf{89.0\%}$ & $\mathbf{83.3}$\err{3.5} & $\mathbf{84.6\%}$ \\
\cmidrule(l){2-6}
 & Teacher                   & $79.3$\err{1.8} & 100\% & $88.4$\err{0.5} & 100\% \\
\midrule
\multirow{5}{*}
{\shortstack[l]{GPT-5.6 Terra\\ + Skill}}
 & Qwen3.6-35B-A3B          & $28.0$\err{1.4} & --- & $59.0$\err{0.7} & --- \\
 & \quad + D-Skill          & $\mathbf{73.1}$\err{3.4} & $\mathbf{89.3\%}$ & $\mathbf{83.1}$\err{1.4} & $\mathbf{93.8\%}$ \\
 & GPT-5.6 Luna             & $25.3$\err{1.1} & --- & $55.2$\err{4.2} & --- \\
 & \quad + D-Skill          & $\mathbf{69.7}$\err{0.8} & $\mathbf{83.4\%}$ & $\mathbf{83.1}$\err{3.2} & $\mathbf{94.6\%}$ \\
\cmidrule(l){2-6}
 & Teacher                   & $78.5$\err{0.7} & $100\%$ & $84.7$\err{1.1} & $100\%$ \\
\bottomrule
\end{tabular}
\end{subtable}

\vspace{0.8em}

\begin{subtable}{\linewidth}
\centering
\subcaption{Chess puzzles}
\label{tab:main-b}
\setlength{\tabcolsep}{3.0pt}
\begin{tabular}{llccc}
\toprule
Teacher & Student config & Score & Est.\ ELO & \%GC \\
\midrule
\multirow{4}{*}{\shortstack[l]{Sonnet 5}}
 & Haiku 4.5\textsuperscript{$\dagger$}    & $41.2$\err{1.1} & $1058$\err{25} & --- \\
 & \quad + D-Skill     & $\mathbf{52.5}$\err{1.1} & $\mathbf{1286}$\err{21} & $\mathbf{75.3\%}$ \\
\cmidrule(l){2-5}
 & Teacher  & $56.2$\err{3.2} & $1383$\err{79} & 100\% \\
\midrule
\multirow{4}{*}{\shortstack[l]{Haiku 4.5}}
 & Qwen3.6-35B-A3B & $24.8$\err{0.4} & $663$\err{74} & --- \\
 & \quad + D-Skill      & $\mathbf{39.0}$\err{3.5} & $\mathbf{979}$\err{72} & $\mathbf{87.7\%}$ \\
\cmidrule(l){2-5}
 & Teacher\textsuperscript{$\dagger$}  & $41.0$\err{1.4} & $1050$\err{26} & 100\% \\
\bottomrule
\end{tabular}
\par\smallskip
\end{subtable}
\end{table}
\begingroup
\renewcommand{\thefootnote}{\fnsymbol{footnote}}%
\footnotetext[2]{The two Haiku~4.5 rows in Table~\ref{tab:main-b} come from different agent harnesses. The Sonnet~5 pair runs in Claude Code; the Haiku~4.5$\rightarrow$Qwen3.6-35B-A3B pair uses the text-tag harness.}%
\endgroup

We evaluate D-Skill across three agentic domains and multiple
teacher--student configurations. For each task and teacher--student
pair, we collect paired rollouts on the same unlabeled tasks,
apply the difference-analysis procedure from Section~\ref{sec:method}, and
consolidate the resulting lessons into a skill document. At evaluation
time, this skill is equipped by the student without modifying its
parameters. 

\paragraph{Datasets and Benchmarks.}
We evaluate D-Skill in three domains with distinct interaction and evaluation
requirements. \textbf{SpreadsheetBench Verified (SSB-V)}
\citep{ma2024spreadsheetbench,shortcut2025ssbverified} evaluates agents on
spreadsheet-editing tasks. We divide its 400 tasks using a fixed shuffled split: 200 tasks form a held-out evaluation set on which we
report accuracy, and a fixed subset of 99 of the remaining 200 tasks forms the distillation set for rollout
collection. \textbf{ALFWorld} \citep{shridhar2021alfworld} evaluates embodied
decision-making in text-based household environments. We collect teacher
and student traces on 57 games from its training split and report success rate on
the standard 134-game test split.
\textbf{Chess} evaluates tactical reasoning in an agentic interaction
setting. We evaluate on a fixed set of 200 puzzles from the public Lichess
puzzle database and collect distillation traces on 100 additional puzzles
that are disjoint from the evaluation set. We apply a strict solve criterion:
a puzzle is counted as correct only when the agent successfully executes the
complete tactical sequence. A more detailed setup for the chess puzzle environment can be found in Appendix \ref{app:chess_implementation}.
Across all three domains, the distillation tasks
provide only task descriptions and interaction traces; no correctness labels
or reference solutions are supplied to the difference-analysis or skill-update stages.

\paragraph{Models and conditions.}
For SpreadsheetBench-Verified and ALFWorld, we consider two black-box teacher
configurations: DeepSeek-V4-Pro-0423 \citep{deepseek2026v4} and GPT-5.6 Terra
\citep{openai2026gpt56}, each equipped with an existing skill solely to strengthen the teacher and create a meaningful teacher--student capability gap.
Before the teacher trajectories are passed to the difference analyst, we remove the teacher's reasoning and all skill- or system-prompt-related content; the analyst receives only the task description and the teacher's externally observable action--observation sequence, so the skill text itself never reaches the analyst. D-Skill therefore treats the teacher as a black-box policy and has no access to the internal mechanism that produces its behavior. From the perspective of the distillation pipeline, equipping the teacher with a skill is operationally equivalent to selecting a stronger black-box teacher, as only the resulting behavior is exposed. 
The students are Qwen3.6-35B-A3B \citep{qwen2026qwen36} and GPT-5.6 Luna \citep{openai2026gpt56}, representing different model families. 
For each teacher--student pair, we report three conditions: the student baseline, the same student augmented with the distilled skill, and the teacher. Both the analyst and consolidation agents are DeepSeek-V4-Pro-0423. For chess puzzles, where the performance gaps among
the GPT-series models are comparatively small, we instead consider two
teacher--student pairs: (a) Sonnet-5 \citep{anthropic2026sonnet5} as the teacher and Haiku-4.5 \citep{anthropic2025haiku45} as the student, and (b) Haiku-4.5 as the
teacher and Qwen3.6-35B-A3B as the student. Sonnet-5 serves as both the analyst and consolidation agent for the chess experiments.

\paragraph{Metrics.}
SSB-V accuracy, ALFWorld success rate, and chess solve rate are reported on a 0--100 scale. We also report the fraction of the corresponding teacher--student gap closed, as defined in Section~\ref{sec:method}. For chess, solve performance is converted to an estimated human-scale tactical rating using the logistic fit described in Appendix~\ref{app:chess_implementation}. Student and teacher results are mean$\pm$standard deviation over two evaluation passes; D-Skill results are mean$\pm$standard deviation over six evaluations, comprising two passes for each of three independently distilled skills.

\subsection{Main results}
\label{main_result}
Table~\ref{tab:main} reports the student’s baseline performance, its performance with the distilled D-Skill documents, and the performance of the teacher that generated the source traces. Across the two agentic benchmarks in Table~\ref{tab:main-a}, D-Skill closes 83--95\% of the teacher--student gap in all eight teacher--student--task configurations without updating the student’s weights. With DeepSeek-V4-Pro as the teacher, Qwen3.6-35B-A3B improves from $28.0$ to $71.7$\err{3.6} on SpreadsheetBench-Verified, while GPT-5.6 Luna improves from $25.3$ to $73.3$\err{1.0}, closing 85\% and 89\% of the gap to the teacher’s score of $79.3$, respectively. On ALFWorld, the same students improve from $59.0$ and $55.2$ to $84.2$\err{3.3} and $83.3$\err{3.5}, closing 85--86\% of the gap. Replacing DeepSeek-V4-Pro with GPT-5.6 Terra, a teacher from a different vendor with a different trace style, yields similar results: Qwen and Luna reach $73.1$\err{3.4} and $69.7$\err{0.8} on SpreadsheetBench-Verified, corresponding to 89\% and 83\% gap closure, and both reach $83.1$ on ALFWorld, corresponding to 94--95\% gap closure. 

Table~\ref{tab:main-b} extends the evaluation to chess puzzles, a tool-mediated, non-language domain with a smaller teacher--student gap. Distilling Sonnet~5 into Haiku~4.5 raises the student’s score from $41.2$ to $52.5$, an 11.3-point improvement equivalent to approximately $228$ Elo and 75.3\% gap closure. Distilling Haiku~4.5 into Qwen3.6-35B-A3B raises the score from $24.8$ to $39.0$, a 14.2-point improvement equivalent to $316$ Elo and 88\% gap closure. These results show that D-Skill generalizes beyond agentic tasks to reasoning-intensive domains such as chess, where success requires strategic planning and sequential decision-making. Representative distilled skills are logged in appendix \ref{app:skills}.

\subsection{Ablations}
\paragraph{Teacher-student contrast improves transfer.}
Table~\ref{tab:ssb-ablation} tests whether D-Skill's gains can be explained by either side of the paired data alone. Across all three conditions, we hold fixed the 99-task distillation set, analyst and consolidation models, lesson format, and 
\begin{wraptable}{r}{0.52\textwidth}
\centering
\captionsetup{skip=4pt}
\vspace{-0\intextsep}
\caption{\textbf{Teacher-student contrast improves performances.} Results average 3 independent skills, each evaluated twice.}
\label{tab:ssb-ablation}
\small
\setlength{\tabcolsep}{2.5pt}
\renewcommand{\arraystretch}{1.1}
\begin{tabular}{@{}lcccc@{}}
\toprule
& \multicolumn{2}{c}{\makecell{Qwen3.6\\35B-A3B}}
& \multicolumn{2}{c}{\makecell{GPT-5.6\\Luna}} \\
\cmidrule(lr){2-3}\cmidrule(lr){4-5}
Method & Score & \%GC & Score & \%GC \\
\midrule
Student trace only
& $47.7$\err{6.7} & $38.3\%$
& $49.8$\err{8.8} & $45.4\%$ \\
Teacher trace only
& $56.3$\err{6.1} & $55.2\%$
& $63.5$\err{5.4} & $70.7\%$ \\
\textbf{D-Skill}
& $\mathbf{71.7}$\err{3.6} & $\mathbf{85.1\%}$
& $\mathbf{73.3}$\err{1.0} & $\mathbf{89.0\%}$ \\
\bottomrule
\end{tabular}
\end{wraptable}
consolidation procedure, varying only which traces are available to the analyst. 
The student-only condition allows the analyst to diagnose weaknesses from the student's 
behavior without access to a stronger policy, whereas the teacher-only condition extracts general guidance from teacher behavior without observing the target student's failures. 
Both one-sided variants improve substantially over the student floor, but neither accounts for the full D-Skill gain. 
For Qwen3.6-35B-A3B, the student-only and teacher-only variants reach $47.7$ and $56.3$, 
respectively, compared with $71.7$ for D-Skill; for GPT-5.6 Luna, they reach $49.8$ and $63.5$, compared with $73.3$.
Full paired contrast therefore closes $85$--$89\%$ of the teacher--student gap, versus $38$--$45\%$ for student-only analysis and $55$--$71\%$ for teacher-only analysis. 
These results indicate that a strong analyst can derive useful guidance from either trajectory in isolation, while directly contrasting matched teacher and student behavior yields larger and more consistent gains by targeting student-specific gaps.

\paragraph{Analyst and consolidation agents.}
\begin{wraptable}{r}{0.45\linewidth}
  \centering
  \small
  \setlength{\tabcolsep}{5pt}
  \vspace{-1.0\intextsep}
  \caption{\textbf{A good analyst is needed.} Chess accuracy of the Qwen student with the skill produced
  by each analyst-consolidation pair. Results average 3 independent skills, each evaluated twice.}
  \label{tab:pipeline-llm}
  \begin{tabular}{l cc}
    \toprule
    & \multicolumn{2}{c}{\textbf{Consolidation}} \\
    \cmidrule(lr){2-3}
    \textbf{Analyst} & Sonnet & Qwen \\
    \midrule
    Sonnet & $\mathbf{39.0}$\err{3.5} \textbf{(88\%)} & $34.3$\err{3.4} (59\%) \\
    Qwen   & $31.1$\err{4.5} (39\%) & $32.6$\err{3.9} (48\%) \\
    \bottomrule
  \end{tabular}
\end{wraptable}
D-Skill uses an LLM in two places: the \emph{analyst}, which contrasts
teacher and student traces and extracts candidate lessons, and the
\emph{consolidation agent}, which merges those lessons into the skill document.
Table~\ref{tab:pipeline-llm} replaces each stage with the student model
itself (Qwen3.6-35B-A3B) on chess, where Haiku~4.5 is the teacher. 
Keeping a strong analyst and letting the student consolidate costs about five
points (39.0 to 34.3), and the resulting document still closes 59\% of
the teacher--student gap. Replacing the analyst is far more damaging:
with a Qwen analyst the document closes only 39--48\% of the gap
regardless of who consolidates. The student analyst produces agreement and divergence items that contain fewer confirmed lessons and noisier gap explanations, and the student consolidation agent fails to compress useful information in the items, resulting in a much longer skill document. The practical implication is that a teacher-grade model is needed only for the analysis stage, which runs once per training batch, and the merging can be delegated to a cheaper model without collapsing the method.

\paragraph{Other skill backends.}
\label{backends}

\begin{wraptable}{r}{0.47\textwidth}
\centering
\vspace{-1.0\intextsep}
\caption{\textbf{D-Skill is compatible with an alternative skill-update backend.} Results average three independently distilled skills, each evaluated twice. The SkillOpt-backed variant retains SkillOpt's held-out validation gate.}
\label{tab:chess-backend}
\small
\setlength{\tabcolsep}{4pt}
\renewcommand{\arraystretch}{1.1}
\begin{tabular}{@{}lcc@{}}
\toprule
Method & Score & \%GC \\
\midrule
Student (Qwen3.6-35B-A3B)         & $24.8$\err{0.4}  & ---   \\
\midrule
SkillOpt                          & $29.6$\err{1.4}  & $29.7\%$ \\
\textbf{D-Skill w/ SkillOpt backend }      & $\mathbf{36.9}$\err{5.7}  & $\mathbf{74.9\%}$ \\
\textbf{D-Skill}                  & $\mathbf{39.0}$\err{3.5} & $\mathbf{87.7\%}$ \\
\bottomrule
\end{tabular}
\end{wraptable}

D-Skill separates difference analysis from skill consolidation: its analyst produces structured lessons rather than a completed document, allowing those lessons to be consumed by any skill-update backend that accepts textual edits. We test this modularity with SkillOpt~\citep{yang2026skillopt}. SkillOpt ordinarily rolls out the student, reflects on unsuccessful trajectories, merges and ranks proposed edits under a fixed budget, and accepts an edit only when it improves held-out validation accuracy. We modify only the input to its reflection step: instead of receiving student trajectories, the SkillOpt analyst receives the D-Skill lessons associated with each minibatch. All downstream stages, including merging, ranking, and validation gating, remain unchanged. 

Table~\ref{tab:chess-backend} reports chess performance with Qwen3.6-35B-A3B on different skill generation framework and update backend. At the same rollout budget of 1 epoch, the lesson-conditioned backend reaches $36.9 \pm 5.7$, outperforming SkillOpt with its original trajectory-reflection analyst by $7.3$ points and performing within the uncertainty of D-Skill's native consolidator. Thus, the resulting lessons also improve a backend originally designed to learn from self-generated failures.
\section{Analysis}

\subsection{Comparing with Other Skill Learning Pipelines}

\begin{table}
\centering
\caption{SpreadsheetBench-Verified: D-Skill vs. Trace2Skill baselines. All numbers are means over six evaluations, comprising two passes for each of three independently distilled skills.}
\label{tab:other-learning-ablation}
\footnotesize
\setlength{\tabcolsep}{3pt}
\begin{tabularx}{\linewidth}{
    >{\raggedright\arraybackslash}X
    *{4}{>{\centering\arraybackslash}p{0.115\linewidth}}
}
\toprule
& \multicolumn{2}{c}{Qwen3.6-35B-A3B}
& \multicolumn{2}{c}{GPT-5.6 Luna} \\
\cmidrule(lr){2-3}\cmidrule(lr){4-5}
Method & Score & \%GC & Score & \%GC \\
\midrule
Student
    & $28.0$\err{1.4} & ---
    & $25.3$\err{1.1} & --- \\
Teacher (DS-V4-Pro + Skill)
    & $79.3$\err{1.8} & 100\%
    & $79.3$\err{1.8} & 100\% \\
\midrule
Trace2Skill (student trace, student-authored)
    & $33.0$\err{4.2} & $9.7\%$
    & $31.1$\err{1.6} & $10.7\%$ \\
Trace2Skill (teacher trace, teacher-authored)
    & $34.2$\err{2.8} & $12.0\%$
    & $37.0$\err{1.6} & $21.7\%$ \\
Trace2Skill (student trace, teacher-authored)
    & $36.0$\err{1.1} & $15.6\%$
    & $46.2$\err{5.9} & $38.6\%$ \\
\textbf{D-Skill}
    & $\mathbf{71.7}$\err{3.6} & $\mathbf{85.1\%}$
    & $\mathbf{73.3}$\err{1.0} & $\mathbf{89.0\%}$ \\
\bottomrule
\end{tabularx}
\end{table}

Table~\ref{tab:other-learning-ablation} compares D-Skill with Trace2Skill~\citep{ni2026trace2skill}, an existing method that distills a skill document from execution traces. The published Trace2Skill configuration improves upon the skill-free student but closes only $10$--$11\%$ of the student--teacher gap. To strengthen the baseline, we additionally construct two variants that progressively provide the resources available to D-Skill: first replacing the student author with the teacher model, and then replacing the student traces with the teacher's own traces, with both resulting skills applied to the student. The strongest variant, teacher-authored documents constructed from student traces, closes $16$--$39\%$ of the gap. Substituting teacher traces does not improve performance further and instead reduces the scores from $36.0$ to $34.2$ for Qwen3.6-35B-A3B and from $46.2$ to $37.0$ for GPT-5.6 Luna. In contrast, D-Skill reaches scores of $71.7$ and $73.3$, closing $85\%$ and $89\%$ of the gap, respectively.

\subsection{Comparing with Weight Updates}
\label{subsec:sft}
\begin{figure}[t]
    \centering
    \begin{subfigure}[t]{0.49\linewidth}
        \centering
        \includegraphics[width=\linewidth]{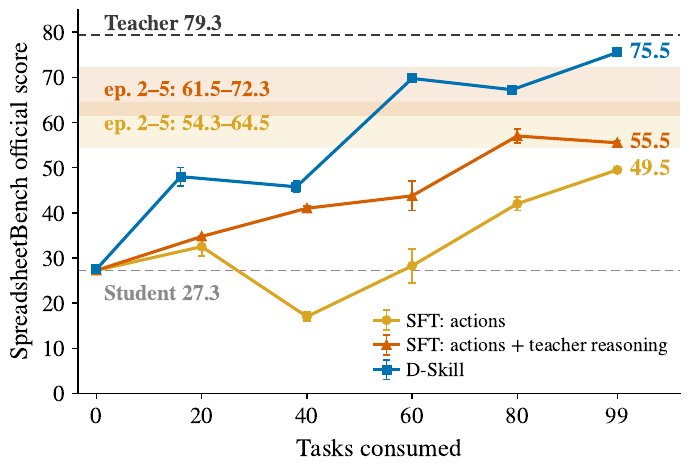}
        \caption{SpreadsheetBench Verified}
        \label{fig:sft-vs-dskill-ssb}
    \end{subfigure}
    \begin{subfigure}[t]{0.49\linewidth}
        \centering
        \includegraphics[width=\linewidth]{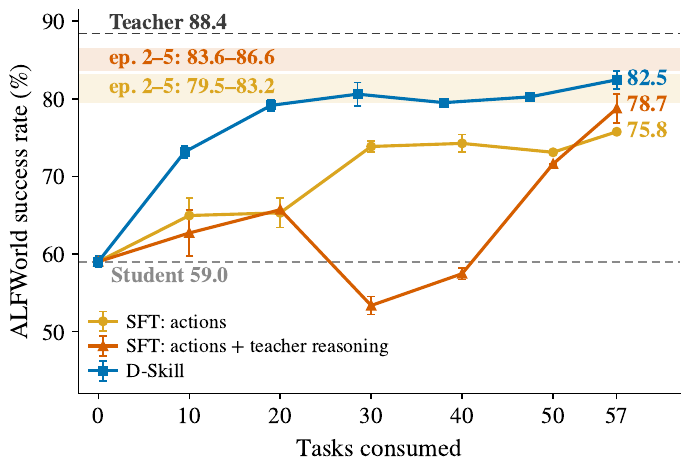}
        \caption{ALFWorld}
        \label{fig:sft-vs-dskill-alf}
    \end{subfigure}
    \caption{\textbf{One-pass D-Skill outperforms one-epoch SFT and remains competitive with its multi-epoch performance.} Both methods use Qwen3.6-35B-A3B as the student and the same teacher-trajectory corpus on (a) SpreadsheetBench Verified and (b) ALFWorld. The x-axis reports the number of teacher tasks processed, and the shaded bands show SFT performance over epochs 2-5.}
    \label{fig:sft-vs-dskill}
\end{figure}

A natural alternative use of the teacher corpus is to fine-tune the student on it. We therefore compare D-Skill with supervised fine-tuning (SFT) using exactly the same DeepSeek-V4-Pro-0423 teacher trajectories. We remove the teacher's skill and system prompt so that both methods receive the same teacher action--observation content. We fine-tune Qwen3.6-35B-A3B with LoRA \citep{hu2022lora} through the Tinker API \citep{thinkingmachines2025tinker} on SpreadsheetBench-Verified and ALFWorld for five epochs under two variants: teacher actions alone and teacher actions together with teacher reasoning. We save checkpoints at several points during the first epoch and at every epoch boundary, evaluating each checkpoint twice. For D-Skill, we evaluate the document after the corresponding consolidation chunk and plot one lineage per benchmark, as independently distilled lineages achieve similar final accuracy. Figure~\ref{fig:sft-vs-dskill} plots performance against the number of unique teacher tasks processed during the first pass over the corpus; the shaded bands show the range attained by SFT over epochs two through five, where the training plateaus. The exact setup of the SFT training can be found in appendix \ref{app:sft}.

D-Skill reaches multi-epoch SFT performance after a single offline pass over the paired traces. On SpreadsheetBench Verified, D-Skill reaches $75.5$ after one pass, compared with $49.5$ for action-only SFT and $55.5$ for reasoning-augmented SFT; after four additional epochs, the two SFT variants reach $54.3$--$64.5$ and $61.5$--$72.3$, respectively. On ALFWorld, D-Skill reaches $82.5$ after one pass, compared with $75.8$ and $78.7$ for the two SFT variants, which rise to $79.5$--$83.2$ and $83.6$--$86.6$ over epochs two through five. 
Thus, using the same teacher-trajectory corpus, D-Skill provides substantially stronger one-pass transfer and remains competitive with multi-epoch SFT without gradient updates or access to the student's weights.



\subsection{Cost Analysis}
\label{subsec:cost}

\begin{wrapfigure}{r}{0.55\linewidth}
  \centering
  \vspace{-1.0\intextsep}
  \includegraphics[width=\linewidth]{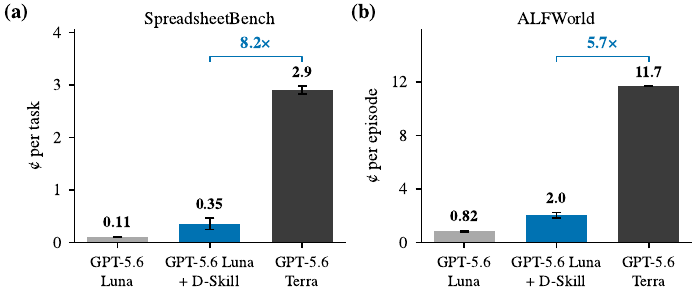}
\caption{\textbf{Inference cost per task at model list prices} on (a) SpreadsheetBench Verified and (b) ALFWorld. Costs are reported in cents episode.}
  \label{fig:cost}
  \vspace{-10pt}
\end{wrapfigure}

Figure~\ref{fig:cost} compares inference cost per task at list prices for models from OpenAI. D-Skill approaches teacher performance at a fraction of the inference cost: GPT-5.6 Luna with the distilled skill is $8.3\times$ cheaper than the teacher on SpreadsheetBench and $5.7\times$ cheaper on ALFWorld, costing only $0.35$ cents per task for SpreadsheetBench and $2.0$ cents per episode for ALFWorld, respectively. At these costs, D-Skill closes 83.4\% of the teacher--student gap on SpreadsheetBench and 94.6\% on ALFWorld. 
However, D-Skill is slightly more expensive than the bare student, as the additional cost comes primarily from reading the skill document and, in some cases, from the extra interaction steps the document induces while solving the task. Overall, D-Skill recovers most of the teacher’s performance while retaining a substantial inference-cost advantage.

\begin{wraptable}{r}{0.36\linewidth}
\vspace{-20pt}
\centering
\caption{\textbf{One-time distillation cost} GPT-5.6 Terra$\rightarrow$Luna, billed at list prices. Per task is per training task.}
\small
\setlength{\tabcolsep}{4pt}
\begin{tabular}{lrrr}
\toprule
 & Tasks & Total & Per task \\
\midrule
SSB-V    & $99$ & \$$3.28$ & $3.3$\textcent \\
ALFWorld & $57$  & \$$1.51$ & $2.6$\textcent \\
\bottomrule
\end{tabular}
\label{tab:distill-cost}
\vspace{-10pt}
\end{wraptable}
Distillation adds a one-time cost that grows linearly with the number of training tasks, since the analyst makes one call per teacher--student trace pair and the consolidator processes the resulting lessons in fixed-size chunks. It costs about $3.3$ cents per training task on SpreadsheetBench and $2.6$ cents on ALFWorld, i.e., \$$3.28$ for our $99$ SpreadsheetBench tasks and \$$1.51$ for our $57$ ALFWorld tasks.
\section{Limitations}
\label{sec:limitations}


D-Skill has three main limitations. First, it requires a capable analyst at distillation time: replacing the analyst with the weaker student substantially reduces the fraction of the gap closed (Table~\ref{tab:pipeline-llm}), although this is a one-time rather than per-inference cost. Second, loading the distilled document increases inference cost relative to the bare student by adding input tokens and, in some cases, inducing additional interaction steps, even though the resulting agent remains substantially cheaper than the teacher. Third, D-Skill can transfer only capabilities that can be expressed as instructions and executed by the student. It can improve procedures such as planning, verification, and tool use, but cannot supply fundamental capabilities the student lacks, such as deeper calculation, perception, or sufficient underlying model capacity.
\section{Conclusion}
\label{sec:conclusion}


We introduced D-Skill, a method that distills the behavioral gap between a black-box teacher and student into a natural-language skill by contrasting paired rollouts on the same tasks. Once the trajectories are collected, D-Skill operates fully offline, requiring no further policy queries, environment interaction, labels, rewards, evaluation feedback, or weight updates. Across SpreadsheetBench Verified, ALFWorld, and chess puzzles, D-Skill closes $75$--$95\%$ of the teacher--student performance gap across multiple model families while retaining a substantial inference-cost advantage over the teacher. 
D-Skill also provides stronger one-pass transfer and remains competitive with multi-epoch SFT when using the same teacher-trajectory corpus, without gradient updates or access to the student's weights. Together, these results demonstrate that natural-language skill distillation is a practical, interpretable, and inference-efficient alternative to weight-based capability transfer in the evaluated domains.
\section{Acknowledgment}
\label{sec:ack}

We would like to thank Fangzhou Zhao, Shiyi Cao, Yichuan Wang, Qiuyang Mang, Zirui Wang, and Hanchen Li for insightful discussions about this work. This work is also supported by Thinking Machine Lab through the Tinker Research Grant. Sky Computing Lab is supported by gifts from Accenture, Algorithmic SuperIntelligence Labs, Amazon, AMD, Anyscale, Broadcom, cmpnd, Google, IBM, Intel, Intesa Sanpaolo, Lambda, Lightspeed, Mirendil, NVIDIA, Samsung SDS, and VESSL.


\newpage
\bibliography{references}
\bibliographystyle{iclr2027_conference}

\newpage
\appendix
\section{Chess Puzzle Environment Setup}
\label{app:chess_implementation}

\begin{table}[h]
\centering
\caption{Analysis tools available to the chess agent. Every tool is a static
computation on a copy of the board; nothing is searched or evaluated.}
\label{tab:chess-tools}
\begin{tabularx}{\linewidth}{@{}l>{\raggedright\arraybackslash}X@{}}
\toprule
Tool & Input and output \\
\midrule
\texttt{<hanging></hanging>} & No input. Lists every piece that is attacked and has no defender, for both sides. \\
\midrule
\texttt{<safety>m1,\,\dots</safety>} & Up to five candidate moves. For each: whether it gives check (checkmate and stalemate named explicitly), the number of legal replies, every forcing reply (checks and captures with the captured piece named), attackers vs.\ defenders of the destination square, and the agent's pieces left undefended afterwards. \\
\midrule
\texttt{<threats></threats>} & No input. Null-move audit: the opponent's forcing moves if the agent passed, and any immediate mate threats. \\
\midrule
\texttt{<board>move</board>} & One move. The board after the agent's move only, with material, undefended pieces, and check/checkmate status. \\
\midrule
\texttt{<simulate>move,\,r1,\,\dots</simulate>} & One candidate and up to five opponent replies, each simulated independently from the position after the candidate: resulting board, material, undefended pieces, and whether the agent is in check. \\
\midrule
\texttt{<line>m1,\,r1,\,m2,\,\dots</line>} & A complete line of two to six plies supplied by the agent (own moves and opponent replies alternating): the final board, material, undefended pieces, and check/mate status. \\
\bottomrule
\end{tabularx}
\end{table}

\paragraph{Task.}
Each episode is a single Lichess tactics puzzle drawn from the public Lichess puzzle
database~\citep{lichess_puzzles}. A puzzle specifies a starting position (FEN), a
scripted opponent move that creates the tactic, and a unique solution line of one or
two solver moves; a two-move puzzle also scripts the opponent's forced reply after the
first solver move. The agent plays the solver side and must produce every solver move
in order. Following Lichess grading, a move is correct if it matches the solution or
delivers checkmate; the first incorrect move ends the puzzle. A puzzle counts as
solved only if all solver moves are correct.

\paragraph{Puzzle sets.}
We filter the database to puzzles with popularity above 70, more than 200 plays,
rating deviation at most 85, and a solution of exactly one or two solver moves.
The \textbf{evaluation set} has 200 puzzles, 50 from each 400-point rating band between
400 and 2000, ratings ranging from 649--1987; 179 two-move and 21 one-move puzzles. The \textbf{distillation set} has 100 puzzles disjoint from the
evaluation set, ratings ranging from
612--2357; 78 two-move and 22 one-move. Teacher and student traces used for
distillation are collected on all 100 distillation puzzles. Baselines that need a
validation split (SkillOpt) divide the same 100 puzzles into 60 training and 40
validation puzzles.

\paragraph{Observation.}
At every solver move the agent receives a fresh user message containing the FEN, an
ASCII board (uppercase = White, rank 8 at the top), the side to move, a side-relative
material count (Q=9, R=5, B=N=3, P=1), the last six moves labeled ``you'' /
``opponent'', the halfmove clock, the complete list of legal moves in SAN, the
agent's optional plan note from its previous move, and the sentence ``This is a
tactics puzzle: there is a clearly best move in this position. Find it.'' In a
two-move puzzle the second observation additionally notes that earlier tool results
referred to the previous position.

\paragraph{Actions and tool protocol.}
The system prompt instructs the agent to reply each turn with either exactly one
tool tag or its committed move wrapped as \texttt{<action>Nf3</action>} (SAN or UCI),
never both. Per solver move the agent may issue up to $K=10$ tool calls; an exact
repeat of an earlier query in the same move is served from a cache at no cost, and
every other call, including malformed or illegal ones, costs one unit. When five or
fewer calls remain, the tool result carries a short low-budget notice, and when the
budget is exhausted the agent is told to commit. A reply that contains neither a tool
tag nor an action tag triggers a corrective message, at most twice per move, asking for
one tool tag or a move; a third tagless reply is scored as no move. A committed move
outside the legal list is likewise scored as incorrect.

The six tools (Table~\ref{tab:chess-tools}) are implemented in-process with
\texttt{python-chess} on a copy of the live board. They only report static facts
about positions the agent specifies; there is no engine, search, or evaluation
anywhere in the environment. Multi-argument tools accept at most five candidate moves
(\texttt{safety}), five opponent replies (\texttt{simulate}), or six plies
(\texttt{line}); \texttt{threats} is unavailable while the agent is in check.

\paragraph{Estimated Elo.}
Lichess rates puzzles on the same Glicko-2 scale as players~\citep{glickman2012glicko2,lichess_puzzles},
so a puzzle rated $r$ is one that a player rated $r$ solves about half the time. For
each evaluation pass we fit a logistic model of solve probability against puzzle
rating, $P(\text{solved}\mid r)=\sigma\!\left(\alpha+\beta\,\tfrac{r-1200}{400}\right)$,
to the 200 evaluation puzzles by maximum likelihood, and report the rating at which
the fitted probability crosses one half, $\widehat{R}=1200-400\,\alpha/\beta$. The
parameterization mirrors the Elo expected-score formula~\citep{elo1978rating} with a
free slope. Each document's Elo is the mean over its two passes; the table reports
mean $\pm$ s.d.\ over distillation replicates for +D-Skill rows and over passes for
floors and teachers.

\section{Difference Analyst Prompt}
\label{app:analyst-prompts}

\newcommand{\slot}[1]{\texttt{<{}<#1>{}>}}

The difference analyst uses one system prompt template for all three domains (below).
Only the slots differ across domains:
\slot{ENVIRONMENT} describes the environment and its action protocol;
\slot{COMPARISON} lists example behavioral dimensions to compare
(e.g., inspection, search, tool use, recovery, completion discipline);
\slot{UNIT} and \slot{EVIDENCE} name the trace unit to quote (steps, or turns and tool calls);
\slot{CHECK} and \slot{GENERALIZE} specify what a runtime check and a generalizable
instruction look like in the domain. In chess, \slot{JUDGE} additionally forbids asserting
that any specific move was correct, and ``agent'' reads ``player''. For
SpreadsheetBench-Verified, \slot{COMPARISON} also asks the analyst to diff the two
agents' produced answer ranges and trace each difference back to the decision that
caused it, without stating which value is correct.

The user message contains the task instruction, the expert trace, and the student trace.
For SpreadsheetBench-Verified it also includes the graded answer range and the cell
contents each agent left in that range (capped at 4{,}000 characters). Neither
reveals the gold values. Each trace is truncated to its first 10{,}000 and last
35{,}000 characters, so the analyst sees the task statement,
the early inspection turns, and the end of every episode.


\lstdefinestyle{prompt}{
  basicstyle=\normalsize\ttfamily,
  columns=fullflexible,
  keepspaces=true,
  breaklines=true,
  breakindent=0pt,
  frame=single,
  framesep=4pt,
  framerule=0.4pt,
  xleftmargin=5pt,
  xrightmargin=5pt,
  framexleftmargin=0pt,
  framexrightmargin=0pt,
  aboveskip=4pt,
  belowskip=8pt
}

\begin{lstlisting}[style=prompt]
<<ENVIRONMENT>>

- The EXPERT trace comes from a much stronger agent whose working style the student should learn to imitate.
- The STUDENT trace comes from the weaker agent you are coaching.

You have NO answer key and NO correctness information for either trace, and you must not invent any. Do not judge whether either agent solved the task<<JUDGE>>. Your job is purely behavioral: identify how the expert's way of working differs from the student's, and convert the most consequential differences into reusable lessons that would make the student's future traces resemble the expert's.

<<COMPARISON>>

Anchor every observation in specific <<UNIT>> visible in the traces -- quote or paraphrase the concrete <<EVIDENCE>>. Never speculate about <<UNIT>> that are not shown.

Write AT MOST 3 items total, ordered by expected impact on the student's future behavior:
- "Failure Memory Item" -- a behavioral gap: what the student did, what the expert did instead, and ONE concrete reusable instruction.
- at most 1 "Success Memory Item" -- a behavior the student already shares with the expert and should keep.

Every item's Content must additionally include one concrete check the student can <<CHECK>>.

Every instruction must generalize across <<GENERALIZE>>.

Output format (exactly this structure, failure items first):

# Failure Memory Item 1
## Title
<short name>
## Description
<the observed student-vs-expert difference, with trace evidence>
## Content
<the single reusable instruction the student should follow>

# Success Memory Item 1
## Title
<short name>
## Description
<the shared behavior, with trace evidence>
## Content
<the instruction to keep doing it>
\end{lstlisting}

\section{Supervised Fine-Tuning Baselines}
\label{app:sft}

The SFT curves in Figure~\ref{fig:sft-vs-dskill} fine-tune the same student on the same
teacher episodes that D-Skill distils, and score every checkpoint with the same
harness and scorer as the rest of the paper.

LoRA rank 16 through the Tinker API~\citep{thinkingmachines2025tinker}; peak learning rate
$10^{-4}$, 5\% warm-up, linear decay; five epochs; loss on assistant tokens
only. Training is ground-truth-free: all teacher episodes are used, failures
included, with no correctness labels. The teacher's system prompt (which
carries its skill package) is replaced by the student's harness prompt, so
training inputs match evaluation inputs. Two target variants: \emph{actions
only}, and \emph{teacher reasoning + actions} with the reasoning in the
student's \texttt{<think>} channel. Checkpoints are saved inside epoch~1 after
a fixed number of training episodes (the $x$-axis of Figure~\ref{fig:sft-vs-dskill}) and
at every epoch end, served in bf16 with thinking enabled (temperature 1.0,
top-$p$ 0.95, top-$k$ 20), and scored as the mean of two passes.

\begin{table}[h]
\centering
\caption{SFT within epoch~1}
\footnotesize
\setlength{\tabcolsep}{4pt}
\begin{tabular}{llcccccc}
\toprule
& & \multicolumn{6}{c}{episodes consumed} \\
\cmidrule(lr){3-8}
& Targets & 20 & 40 & 60 & 80 & 99 & \\
\midrule
\multirow{2}{*}{SSB} & actions & $32.5$\err{2.8} & $17.0$\err{1.4} & $28.3$\err{5.3} & $42.0$\err{2.1} & $49.5$\err{0.7} & \\
 & reasoning + actions & $34.8$\err{0.4} & $41.0$\err{0.7} & $43.8$\err{4.6} & $57.0$\err{2.1} & $55.5$\err{0.7} & \\
\midrule
& & 10 & 20 & 30 & 40 & 50 & 57 \\
\cmidrule(lr){3-8}
\multirow{2}{*}{ALFWorld} & actions & $65.0$\err{3.2} & $65.3$\err{2.7} & $73.8$\err{1.1} & $74.2$\err{1.6} & $73.1$\err{0.0} & $75.8$\err{0.5} \\
 & reasoning + actions & $62.7$\err{4.2} & $65.7$\err{0.0} & $53.4$\err{1.6} & $57.5$\err{1.1} & $71.6$\err{0.0} & $78.8$\err{2.6} \\
\bottomrule
\end{tabular}
\label{tab:sft-epoch1}
\end{table}

\begin{table}[h]
\centering
\caption{SFT across epochs}
\footnotesize
\setlength{\tabcolsep}{5pt}
\begin{tabular}{llcccc}
\toprule
& Targets & epoch 2 & epoch 3 & epoch 4 & epoch 5 \\
\midrule
\multirow{2}{*}{SSB} & actions & $54.3$\err{2.5} & $58.8$\err{5.3} & $64.5$\err{2.1} & $63.8$\err{1.8} \\
 & reasoning + actions & $61.5$\err{0.7} & $72.3$\err{0.4} & $70.8$\err{0.4} & $71.5$\err{1.4} \\
\midrule
\multirow{2}{*}{ALFWorld} & actions & $79.5$\err{1.6} & $82.4$\err{0.5} & $82.0$\err{1.1} & $83.2$\err{0.6} \\
 & reasoning + actions & $83.5$\err{1.1} & $86.2$\err{1.6} & $85.8$\err{1.1} & $86.6$\err{2.1} \\
\bottomrule
\end{tabular}
\label{tab:sft-epochs}
\end{table}

\section{Distilled Skill Documents}
\label{app:skills}

For each benchmark we show excerpts of the highest-scoring D-Skill document with Qwen3.6-35B-A3B as the student. By a fixed rule, each excerpt is the document's opening section and the first three rules of its \emph{Common Pitfalls} section, rendered from Markdown; [\ldots] marks omitted text. The complete documents are in the supplementary material.

\subsection*{SpreadsheetBench-Verified}
\begin{quote}\sloppy
\noindent\textbf{Core Workflow}
\par\noindent\hangindent=1.2em\hangafter=1\makebox[1.2em][l]{1.}Inspect the entire workbook: load and print all non-empty cells with coordinates, values, types, and number formats.
\par\noindent Always follow a two-stage loop: inspect the entire workbook before any modification, and verify the output after saving by reloading the workbook from disk (do not rely on in-memory state). This habit catches misplacements and provides a clear picture of the result.
\par\noindent\hangindent=1.2em\hangafter=1\makebox[1.2em][l]{\textbullet}[!] CRITICAL: Never create a new workbook with \texttt{openpyxl.\allowbreak{}Workbook(\allowbreak{})}. Always load the existing file with \texttt{openpyxl.\allowbreak{}load\_\allowbreak{}workbook(\allowbreak{}spreadsheet\_\allowbreak{}path)} and modify it in place. After saving, reload the output and print the exact answer range (as stated in \texttt{answer\_\allowbreak{}position}) to confirm all values are plain data (no formula strings) and that all original sheet names are still present. Only call \texttt{TASK\_\allowbreak{}COMPLETE} after this check passes.
\par\noindent After this inspection, you have all the information needed to write the solution; do not inspect again unless you encounter a specific issue during implementation. Proceed directly to writing the script.
\par\noindent\hangindent=1.2em\hangafter=1\makebox[1.2em][l]{2.}After 2--3 focused inspection steps, write a Python script that produces the required output and saves it to the \texttt{output\_\allowbreak{}path} specified in the task. If you have inspected the workbook for more than 3 turns without writing any code, or if you receive a system warning about a missing output file, stop and immediately write the output script, even if you are unsure of some details. Verify the file exists with \texttt{assert os.\allowbreak{}path.\allowbreak{}exists(\allowbreak{}OUTPUT\_\allowbreak{}PATH)}, then reload it with \texttt{data\_\allowbreak{}only=\allowbreak{}True} and run the verification checks (see Verification). Never continue exploring without producing the deliverable.
\par\noindent\hangindent=1.2em\hangafter=1\makebox[1.2em][l]{\textbullet}Always use the exact absolute file paths provided in the task instruction (\texttt{spreadsheet\_\allowbreak{}path} and \texttt{output\_\allowbreak{}path}). Never change directory or guess paths; pass them directly to Python scripts.
\par\noindent\hangindent=1.2em\hangafter=1\makebox[1.2em][l]{\textbullet}If the task allows solving without VBA, always use Python/openpyxl; never write VBA macros. This keeps the solution simple, auditable, and easy to verify.
\par\smallskip\noindent[\ldots]
\par\medskip\noindent\textbf{Common Pitfalls}
\par\noindent\hangindent=1.2em\hangafter=1\makebox[1.2em][l]{\textbullet}Writing a formula string (e.g., \texttt{=\allowbreak{}VLOOKUP(\allowbreak{}.\allowbreak{}.\allowbreak{}.\allowbreak{})}) instead of the computed result. The grader expects the evaluated value, not the formula text. Always compute programmatically and write the result.
\par\noindent\hangindent=1.2em\hangafter=1\makebox[1.2em][l]{\textbullet}\textbf{String matching}: Always check \texttt{isinstance(\allowbreak{}value,\allowbreak{} str)} before performing string operations. Normalize both the lookup value and cell text by stripping whitespace and case-folding (e.g., \texttt{.\allowbreak{}strip(\allowbreak{}).\allowbreak{}casefold(\allowbreak{})} or \texttt{.\allowbreak{}strip(\allowbreak{}).\allowbreak{}upper(\allowbreak{})} depending on need). Apply the same normalization when comparing sheet names, header labels, or any string identifiers from different sources. Never convert arbitrary cell types to strings with \texttt{str(\allowbreak{}.\allowbreak{}.\allowbreak{}.\allowbreak{})} for text matching; this can cause false positives or missed deletions.
\par\noindent\hangindent=1.2em\hangafter=1\makebox[1.2em][l]{\textbullet}\textbf{Number formats}: After writing any value that should preserve the source's format (e.g., date, currency, percentage, or any value copied from another cell), explicitly copy \texttt{.\allowbreak{}number\_\allowbreak{}format} from the source cell (\texttt{target.\allowbreak{}number\_\allowbreak{}format =\allowbreak{} source.\allowbreak{}number\_\allowbreak{}format}). Default formats can cause incorrect display.
\par\smallskip\noindent[\ldots]
\end{quote}

\subsection*{ALFWorld}
\begin{quote}\sloppy
\noindent\textbf{Core Approach}
\par\noindent Follow the canonical task flow:
\par\noindent\hangindent=1.2em\hangafter=1\makebox[1.2em][l]{1.}Locate and take the target object.
\par\noindent\hangindent=1.2em\hangafter=1\makebox[1.2em][l]{2.}Navigate to the appropriate processing station.
\par\noindent\hangindent=1.2em\hangafter=1\makebox[1.2em][l]{3.}Use the specific interaction verb that appears in the admissible action list.
\par\noindent\hangindent=1.2em\hangafter=1\makebox[1.2em][l]{4.}Navigate to the goal receptacle.
\par\noindent\hangindent=1.2em\hangafter=1\makebox[1.2em][l]{5.}Place the object there.
\par\noindent Follow the natural subgoal order implied by the instruction: acquire $\rightarrow$ transform $\rightarrow$ deliver.
\par\noindent \textbf{Multi-object rule:} When the task instruction contains multiple object nouns (e.g., ``examine the bowl with the desklamp''), treat every noun as a required object that must be found and used. Do not treat any as a mere locator. Before taking any action, list all object nouns and check off each as you find them. If a required object is missing, search for it before interacting with the others.
\par\noindent Act on current observations and admissible actions immediately. Execute a direct, linear path toward the goal state. Before choosing any action, ask: ``Does my current observation already give me enough information to take a productive step toward the goal, or am I only verifying what I already know?'' If the former, take that step directly; if the latter, skip it.
\par\noindent Trust admissible actions as ground truth. When you arrive at a location and the admissible actions list contains an interaction with your target object (e.g., \texttt{take mug 1 from diningtable 1}), select it without re-verifying from memory or prior observations.
\par\noindent Before acting, scan the admissible action list and match your next step to an available action verb. Do not assume an action is available---verify it appears in the list.
\par\smallskip\noindent\textit{Task Completion}
\par\noindent When at a target surface and holding the object to be placed there, scan the admissible actions for a \texttt{move .\allowbreak{}.\allowbreak{}.\allowbreak{} to .\allowbreak{}.\allowbreak{}.\allowbreak{}} command that names both the held object and the destination you are facing. Select that exact action immediately---do not check inventory, look around, or deliberate. Execute it without delay if the goal requires placement there.
\par\noindent After successfully placing the object at the goal surface, stop acting. The task is complete. Do not reopen containers, check inventory, or retrieve items.
\par\noindent After executing a \texttt{move} or \texttt{put} action that places the target object at the goal destination, read the observation for confirmation (e.g., ``You move the X to the Y''). Once confirmed, stop acting immediately. Do not retrieve the object, reopen containers, or restart any part of the task sequence. Runtime check: ``Did my last observation confirm the target object is now in the target destination?'' If yes, the task is complete.
\par\smallskip\noindent[\ldots]
\par\medskip\noindent\textbf{Common Pitfalls}
\par\noindent\hangindent=1.2em\hangafter=1\makebox[1.2em][l]{\textbullet}\textbf{DO NOT} use \texttt{inventory} or \texttt{look} as a search action when the target is not in hand and not on the current surface. Navigate to the next unvisited candidate surface instead. \texttt{look} is only for reorientation when no plausible \texttt{go to} destination exists; if any \texttt{go to} action leads to an unvisited surface, select it --- never substitute \texttt{look} for navigation.
\par\noindent\hangindent=1.2em\hangafter=1\makebox[1.2em][l]{\textbullet}\textbf{DO NOT} revisit surfaces you have already examined and found empty. Track visited locations and skip them.
\par\noindent\hangindent=1.2em\hangafter=1\makebox[1.2em][l]{\textbullet}\textbf{DO NOT} repeat the same action (e.g., \texttt{examine}, \texttt{look}) on the same surface or object if the previous attempt returned no new information. If an observation does not advance progress toward the goal, immediately switch to a different admissible action: go to an unexplored location or use a different interaction verb. Runtime check: After any observation, ask: ``Did this reveal new information that moves me closer to the goal?'' If not, pick an action you have not already tried on this object or surface.
\par\smallskip\noindent[\ldots]
\end{quote}

\subsection*{Chess puzzles}
\begin{quote}\sloppy
\noindent\textbf{When NOT to Analyze: Fast Paths and Stopping Rules}
\par\noindent Rules in this section license SKIPPING work: when a single cheap check settles the position, when further verification cannot change the decision, and when a line of analysis should be abandoned. Every entry must name (a) the exact trigger that makes the shortcut safe and (b) the condition under which the shortcut must NOT be taken. A rule without both parts does not belong here.
\par\noindent\hangindent=1.2em\hangafter=1\makebox[1.2em][l]{\textbullet}Trigger: the mating/tactical idea's supporting facts (king can't escape, no capture or block available) have already been stated once and confirmed. Shortcut: stop re-verifying and commit to the move. Do NOT take this shortcut if the first check was itself based on a hand-parsed FEN that hasn't been cross-checked against the printed board/legal-move list.
\par\smallskip\noindent[\ldots]
\par\medskip\noindent\textbf{Common Pitfalls}
\par\noindent\hangindent=1.2em\hangafter=1\makebox[1.2em][l]{\textbullet}Restating the same candidate-move analysis in prose a second time with no new tool result. Ask ``did I already write this exact conclusion in a prior turn?'' --- if yes, emit a tool call or \textless{}action\textgreater{} instead of another paragraph.
\par\noindent\hangindent=1.2em\hangafter=1\makebox[1.2em][l]{\textbullet}After a tool call discards or confirms a candidate, the very next turn must contain another tool call or an \textless{}action\textgreater{} tag --- never a turn of pure manual analysis. If any sentence or Q\&A pair (e.g. ``If X, Y plays Z? No.'') appears twice in your own output, that is an infinite loop: stop mid-thought and issue a tool/action tag immediately, do not finish the paragraph first.
\par\noindent\hangindent=1.2em\hangafter=1\makebox[1.2em][l]{\textbullet}Committing to a move that your own analysis already concluded loses material/``is bad'' against its critical reply, or whose every self-calculated line was tagged bad, equal, or drawing. Once your own calculation (with or without a tool) produces a concrete refutation of a candidate, that candidate is dead --- do not loop back and re-examine it again hoping for a different conclusion. Either move to a genuinely different candidate and verify it, or re-run \texttt{\textless{}simulate\textgreater{}}/\texttt{\textless{}safety\textgreater{}} on the exact refuted line to check if the refutation still holds; never restate a plan that contradicts your own just-derived refutation.
\par\smallskip\noindent[\ldots]
\end{quote}

\end{document}